\documentclass[10pt,twocolumn,letterpaper]{article}
\usepackage{cvpr}              % To produce the CAMERA-READY version
\usepackage{graphicx}
\usepackage{amsmath}
\usepackage{amssymb}
\usepackage{booktabs}

\usepackage{multirow}
\usepackage{pifont}
\usepackage{float}
\usepackage{color, soul}

\usepackage[pagebackref,breaklinks,colorlinks]{hyperref}

\usepackage[capitalize]{cleveref}
\crefname{section}{Sec.}{Secs.}
\Crefname{section}{Section}{Sections}
\Crefname{table}{Table}{Tables}
\crefname{table}{Tab.}{Tabs.}

\def\cvprPaperID{5803} % *** Enter the CVPR Paper ID here
\def\confName{CVPR}
\def\confYear{2023}

\begin{document}

\title{Neighbour Consistency Guided Pseudo-Label Refinement for Unsupervised Person Re-Identification}

\author{De Cheng$^{1}$\footnotemark[1], Haichun Tai$^{1}$\footnotemark[1], Nannan Wang$^{1}$\footnotemark[2], Zhen Wang$^{2}$, Xinbo Gao$^{3}$,\\
$^{1}$ Xidian University,
$^{2}$ Zhejiang Lab,
$^{3}$ Chongqing University of Posts and Telecommunications\\
%{\tt\small \{dcheng,nnwang\}@xidian.edu.cn, tongliang.liu@sydney.edu.au, yxning@stu.xidian.edu.cn,}\\
%{\tt\small  bhanml@comp.hkbu.edu.hk, gang.niu.ml@gmail.com,gaoxb@cqupt.edu.cn, sugi@k.u-tokyo.ac.jp}
}

\maketitle

\footnotetext[1]{Equal contribution}
\footnotetext[2]{Corresponding author.}

%%%%%%%%% ABSTRACT
\begin{abstract}
 Unsupervised person re-identification (ReID) aims at learning discriminative identity features for person retrieval without any annotations. Recent advances accomplish this task by leveraging clustering-based pseudo labels, but these pseudo labels are inevitably noisy which deteriorate model performance. In this paper, we propose a Neighbour Consistency guided Pseudo Label Refinement (NCPLR) framework, which can be regarded as a transductive form of label propagation under the assumption that the prediction of each example should be similar to its nearest neighbours'. Specifically, the refined label for each training instance can be obtained by the original clustering result and a weighted ensemble of its neighbours' predictions, with weights determined according to their similarities in the feature space. In addition, we consider the clustering-based unsupervised person ReID as a label-noise learning problem. Then, we proposed an explicit \emph{neighbour consistency regularization} to reduce model susceptibility to over-fitting while improving the training stability. The NCPLR method is simple yet effective, and can be seamlessly integrated into existing clustering-based unsupervised algorithms. Extensive experimental results on five ReID datasets demonstrate the effectiveness of the proposed method, and showing superior performance to state-of-the-art methods by a large margin.
\end{abstract}

%%%%%%%%% BODY TEXT
\section{Introduction}
\label{sec:intro}
Person re-identification (ReID) aims to train a deep model capable of retrieving a person of interest across multiple cameras. This task has attracted increasing attention, due to its great application in video surveillance system. Although supervised methods have achieved impressive performances, they require to annotate large amount of cross-camera labels of the surveillance data, which is labor-intensive, costly, and eventually leads to limited practical application in real-world scenarios.
Therefore, developing effective unsupervised methods for person retrieval from unlabeled data is very appealing and important, not only in academic sector but also for the industrial fields.

Existing state-of-the-art unsupervised learning (USL) ReID methods leverage the pseudo-labels obtained from unsupervised clustering~\cite{dai2021cluster,ge2020self} or $k$-nearest neighbor search~\cite{wang2020unsupervised,lin2020unsupervised} to train the deep model. The training scheme of these methods usually alternates between the following two steps: 1) Generating pseudo-labels for the training examples through some clustering-based methods, e.g., DBSCAN~\cite{ester1996density}; 2) Optimizing the deep neural network with these pseudo labels in a supervised manner by some metric learning objectives, such as triplet loss~\cite{cheng2016person}, InfoNCE loss~\cite{he2020momentum}. Although these pseudo-label-based methods have achieved remarkable performances, there still contains a large performance gap between the purely USL methods and supervised learning methods. The rationale behind this is that the generated pseudo labels inevitably contain a portion of noise, which could significantly deteriorate the model performance due to the model memorization/over-fitting to the noisy labels~\cite{cho2022part,cheng2022instance}. Therefore, how to mitigate the side-effects of the  inaccurate clustering results automatically becomes the key issue for these clustering-based unsupervised methods.

To address this problem, we propose a neighbour consistency guided pseudo-label refinement method (NCPLR) for USL person ReID. Although recent advances in pseudo-label refinement could reduce the label noise to a certain extent, these methods usually adopt multiple predictions from auxiliary backbone networks for mutual confirmation of the estimated pseudo-labels~\cite{ge2020mutual,zhang2021refining}, or employ some additional information (e.g., part-based refinement~\cite{cho2022part}) to improve the quality of the pseudo labels, resulting in high computation costs during model training. Our proposed NCPLR algorithm is simple yet effective, just through assembling the predictions of its nearest neighbours to refine the pseudo label. It can be seamlessly integrated into existing clustering-based USL methods. Even for learning with noisy labels, although neighbours have often been used to identify mislabeled examples~\cite{iscen2022learning}, few works have considered the use of neighbours to generate or refine the pseudo labels.

%To well address the aforementioned challenges, we propose a novel neighbour consistency guided pseudo-label refinement (NCPLR) algorithm for unsupervised person Re-ID. It seeks to transfer labels from its neighbouring instances in the feature space, and encourage each example to have similar predictions to its neighbours. The proposed NCPLR strategy is simple yet effective can be seamlessly integrated into existing clustering-based unsupervised method. Specifically, our proposed NCPLR method is inspired by the label propagation algorithms~\cite{iscen2019label,kipf2016semi,zhou2003learning}, which try to transfer labels from the supervised instances to their neighbouring unsupervised examples based on the similarities in the feature space. As we know, existing methods for label propagation represent the inductive learning where they could produce labels for the given examples during model training, the NCPLR can be regarded as an inductive form of label propagation for pseudo-label self-refinement. Therefore, the refined label for each training example can be obtained by the original cluster result and a weighted combination of its neighbours' predictions, where the weights are determined by their similarity in the feature space. The motivation behind this is that the NCPLR is to enable the incorrect labels among the generated pseudo labels to be improved or at least attenuated by the labels of their neighbours~\cite{iscen2022learning}, relying on the moderate assumption that the prediction of each example should be similar to its neighbours'.

The proposed NCPLR method is inspired by the label propagation algorithms~\cite{iscen2019label,kipf2016semi,zhou2003learning}, which try to transfer labels from the supervised instances to their neighbouring unsupervised examples based on the similarities in the feature space. It seeks to transfer labels from its neighbouring instances, and encourage each example to have similar predictions to its neighbours'. As we know, existing methods for label propagation represent the transductive learning, where they could produce labels for the given examples during model training. NCPLR can also be regarded as an transductive form of label propagation for pseudo-label refinement. Specifically, the refined label for each training example can be obtained by the original clustering result and a weighted combination of its neighbours' predictions, with weights determined according to their similarities in the feature space. The motivation behind this is that the NCPLR is to enable the incorrect labels among the generated pseudo labels to be improved or at least attenuated by the labels of their neighbours~\cite{iscen2022learning}, relying on the moderate assumption that the prediction of each example should be similar to its nearest neighbours'.

In addition, we consider the USL person ReID as part of the label-noise learning problem. For label-noise learning, the key challenge is that the deep model could easily memorize and over-fit to the noisy labels during model training, which leads to severe performance degradation. Therefore, effective measures should also be taken to address the over-fitting problem. In term of this issue, we further leverage the explicit \emph{neighbour consistency regularization} to encourage each example to have similar predictions as their neighbours', and penalize the divergence of each example's prediction from a weighted combination of its neighbours' predictions.
Since the similarity graph is computed in the feature space instead of directly using their predictions, the \emph{neighbour consistency regularization} can be seen as bootstrapping the learned feature representations. This could be more helpful to reduce the susceptibility to over-fitting and improve the training stability, as the last fully connected classification layers are more prone to memorize/overfit to the noisy labels~\cite{liu2020early}.

%Different from existing pseudo-label refinery strategies, our proposed neighbour consistency guided pseudo-label refinement (NCPLR) algorithm seeks to transfer labels from its neighbouring instances in the feature space, and encourage each example to have similar predictions to its neighbours. The proposed NCPLR strategy is simple yet effective can be seamlessly integrated into existing clustering-based unsupervised method.

The main contributions of this paper are as follows,
\begin{itemize}
  \item We propose a neighbour consistency guided pseudo-label refinement algorithm for USL person ReID, which refine the pseudo label through a weighted combination of its neighbours' predictions, with weights determined by their similarities in the feature space.
  \item We consider the clustering-based USL person Re-ID task as the label-noise learning problem, and we further leverage the explicit \emph{neighbour consistency regularization} to reduce the model susceptibility to over-fitting and improve the training stability.
  \item The proposed NCPLR algorithm is simple yet effective, and can be seamlessly integrated into existing clustering-based unsupervised methods. Besides, extensive experimental results demonstrate the effectiveness of the proposed method, and the performances are superior to state-of-the-art methods by a large margin.
\end{itemize}

%-------------------------------------------------------------------------
\section{Related Work}

\begin{figure*}[!t]
\centerline{\includegraphics[width=17.5cm]{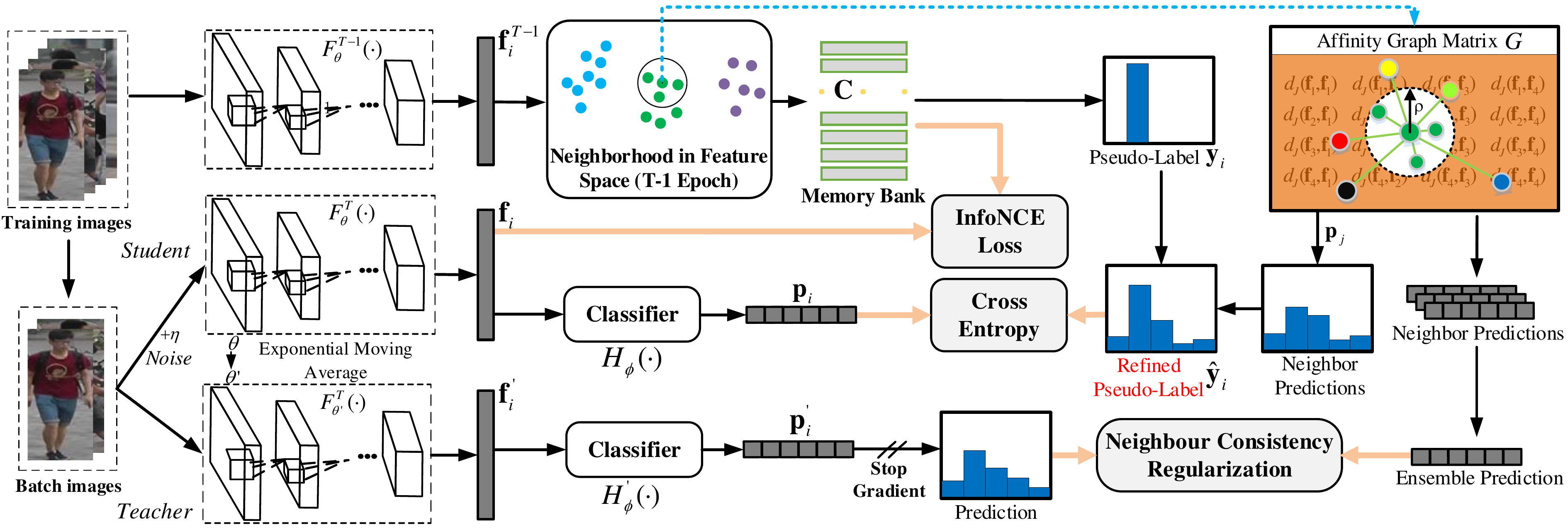}}
\caption{This is the proposed Neighbour Consistency guided Pseudo-Label Refinement (NCPLR) framework for unsupervised person ReID. In the clustering stage, the pseudo labels and overall affinity graph $\textbf{G}$ for dataset $\mathcal{D}$ are generated based on the clustering results. While in the training stage, the pseudo-label refinement is first performed, and then the refined cross-entropy, InfoNCE loss and \emph{neighbour consistency regularization} are adopted to jointly optimize the deep model. }
\label{overallFramework}
\end{figure*}

%\caption{This is the proposed Neighbour Consistency guided Pseudo-Label Refinement (NCPLR) framework for unsupervised person ReID. The framework mainly contains five components: 1) The two-stream mean-teacher network working as the feature extractor; 2) The overall affinity graph $\textbf{G}$ constructed on the dataset $\mathcal{D}$ based on previous epoch features; 3) The memory-bank based InfoNCE loss; 4) The proposed NCPLR module; 5) The \emph{neighbour consistency regularization} term.}

\textbf{Unsupervised Person Re-ID.} Existing unsupervised person Re-ID methods can be roughly divided into two categories: the unsupervised domain adaptation (UDA) and the purely unsupervised learning (USL) person Re-ID, depending on whether or not the labeled source domain data are used. The UDA person Re-ID methods emphasize on transferring knowledge from the labeled source to unlabeled target datasets, through feature distribution alignment~\cite{lin2018multi,wang2018transferable}, or image style transfer~\cite{wei2018person,deng2018image}, and then label propagation~\cite{zhang2021refining,cho2022part}. In this paper, we focus on the purely USL person Re-ID. Current advances for USL Re-ID mainly leverage pseudo label generation by clustering~\cite{dai2021cluster,ge2020self}, as well as some self-supervised techniques~\cite{cheng2022hybrid}.

Representative clustering based pseudo-label generation methods include: PUL~\cite{fan2018unsupervised}, which proposes a progressive USL method by alternating between clustering and fine-tuning; BUC~\cite{lin2019bottom} proposes a bottom-up clustering approach to gradually merge single example into larger clusters; SpCL~\cite{ge2020self} utilizes the self-paced method to gradually generate more reliable clusters; HCT~\cite{zeng2020hierarchical} proposes the hierarchical clustering method to improve the quality of the pseudo labels; ClusterContrast~\cite{dai2021cluster} stores the unique cluster representation and computes contrast loss at cluster level to consistently update the clusters. Some other methods, e.g., CycAs~\cite{wang2020cycas} and TSSL~\cite{wu2020tracklet}, employ the tracklet information to estimate the pseudo labels.
%The representative nearest neighbour search methods include: SSL~\cite{lin2020unsupervised}, which explores to use the pair-wise similarity based softened pseudo-labels to train the deep model;  MultiLabel~\cite{wang2020unsupervised} and SoftMul~\cite{yu2019unsupervised} proposed to use the multi-label reference learning to iteratively boost the Re-ID performances.
%Some other methods, e.g. CycAs~\cite{wang2020cycas} and TSSL~\cite{wu2020tracklet}, employ the tracklet information to estimate the pseudo labels.

\textbf{Label Refinement Method.} The generated pseudo labels used for USL ReID inherently contain a portion of label noise, which could significantly degrade the performances. To tackle this problem, two effective strategies have been proposed: 1) Performing robust clustering on the dataset~\cite{ge2020self,zhai2020ad}; 2) Pseudo-label refinery strategy~\cite{cho2022part,zhang2021refining}. The modified clustering algorithms~\cite{ge2020self} usually define extra criteria to collect more reliable clusters, which struggle to keep balance between the compactness and looseness of clusters. Though such methods demonstrate their effectiveness in some cases, they are always parameter sensitive and highly depended on the instance features themselves.

Representative label refinement algorithms include some nearest neighbour search methods~\cite{lin2020unsupervised,wang2020unsupervised,ge2020mutual} and some additional information assisted pseudo-label refinement methods~\cite{cho2022part,zhang2021refining}. SSL~\cite{lin2020unsupervised} explores to use the pair-wise similarity based softened pseudo-labels to train the deep model;  MultiLabel~\cite{wang2020unsupervised} and SoftMul~\cite{yu2019unsupervised} proposes to use multi-label reference learning to iteratively boost the Re-ID performances. MMT~\cite{ge2020mutual}, MEBNet~\cite{zhang2021refining} and MultiExperts~\cite{zhai2020multiple} proposes to refine the pseudo labels by using multiple predictions from auxiliary networks and train the network in a mutual learning manner. PLCCG~\cite{zhang2021refining} introduces to refine pseudo labels with clustering consensus over training epochs and temporal ensembling techniques. PPLR~\cite{cho2022part} proposes the part-based  pseudo-label refinement to employ additional reliable complementary information to improve the quality of the pseudo labels. Different from these methods that use additional information or networks which may require extra computation cost, our proposed pseudo-label refinement strategy just uses its neighbours' predictions, which is simple yet effective and can be seamlessly integrated into existing clustering based USL person ReId methods.

\textbf{Learning with Noisy Labels.} Due to the difficulties in collecting large-scale high-quality annotated labels in real-world scenarios, training deep models that are robust to label noise are therefore highly attractive. Our USL method is related with existing statistically inconsistent label-noise learning algorithms. These methods usually explore some heuristics to reduce the side-effect of the label noise.

The data cleaning methods aim to select some reliable examples~\cite{cheng2020learning,lyu2019curriculum}. The label correction methods try to improve the quality of labels during model training through label propagation or bootstrapping approaches. For example, JointOpt~\cite{tanaka2018joint} and PENCIL~\cite{yi2019probabilistic} replace the noisy labels with soft or hard pseudo-labels, Bootstrap~\cite{reed2014training} and M-Correction~\cite{arazo2019unsupervised} try to correct the labels by combination of noisy labels and the model predictions in a convex manner, SELC~\cite{lu2022selc} proposes to correct the labels through a progressive self-ensemble strategy. Some other regularization-based methods ~\cite{liu2020early,lu2021confidence} prevent memorization of noisy labels by using a regularizer. Specifically, our method regards the USL person Re-ID as a type of label noise learning problem. Then, we consider the use of neighbours to refine the generated pseudo-labels, and propose the \emph{neighbour consistency regularization} to address the over-fitting problem.

\section{The Proposed Method}

%We propose a Neighbour Consistency guided Pseudo-Label Refinement (NCPLR) framework which exploits to transfer labels from neighbouring examples according to their similarity in the feature space, which is inspire by label propagation methods for semi-supervised learning~\cite{}. Following existing clustering-based unsupervised person ReID methods~\cite{}, our algorithm also alternates between the clustering and training stages.

We propose a Neighbour Consistency guided Pseudo-Label Refinement (NCPLR) framework for USL person ReID, which seeks to transfer labels from neighbouring instances according to their similarities in the feature space. Following existing clustering-based USL person ReID methods~\cite{dai2021cluster,ge2020self}, our algorithm also alternates between the clustering and training stages.

Figure~\ref{overallFramework} illustrates the overall framework of the NCPLR method. In the clustering stage, we first perform clustering over the features extracted by $T-1$ epoch model $F_{\theta}^{T-1}$. Then, we assign pseudo labels to the training examples according to the clustering results, and also initialize the memory bank using the cluster centers. Meanwhile, we need to construct the overall affinity graph $\textbf{G}$ for the whole dataset $\mathcal{D}$ based on the \emph{Jaccard Distance} between instance features. During the training stage, given one training example $\textbf{x}_i$ with two different data augmentations, we can obtain their corresponding features $\textbf{f}_i$ and $\textbf{f}'_i$, as well as the classifier predictions $\textbf{p}_i$ and $\textbf{p}'_i$, based on current two-stream mean-teacher framework, $F_{\theta}$ and $F_{\theta'}$, where the parameters of the teacher network $\theta'$ are updated by the \emph{exponential moving average} strategy from the student network $\theta$. Then we perform pseudo-label refinement in its neighbourhood for each instance, to obtain the refined pseudo label $\hat{\textbf{y}}_i$ and the ensemble prediction. Finally, the refined Cross-Entropy, InfoNCE loss and the proposed \emph{neighbour consistency regularization} are adopted to jointly optimized the deep model.

%We then use the label propagation method for pseudo-label refinement, where the refined  label for each training sample can be obtained by the original cluster result and a weighted combination of its neighbours' predictions. Note that the neighbour relationships between instances are determined by their similarities in the feature space, which is pre-computed before every training epoch over all the training dataset. In the training stage, we use these refined pseudo-labels to train the deep model. In order to prevent the deep model from overfitting to the noisy labels, we further adopt the neighbour consistency regularization through penalizing the divergence of each example's prediction from a weighted combination of its neighbours' predictions. The features extracted by the trained model are then used in the next clustering stage to update the pseudo-labels. The overall framework is illustrated in Figure~\ref{}.

\subsection{Problem Formulation}
Given a person ReID dataset denoted as $\mathcal{D}=\{\textbf{x}_i\}_{i=1}^{N}$, where $\textbf{x}_i$ represents the $i$-th training example, and $N$ is the total number of training examples. The goal of USL person ReID is to train a robust model on $\mathcal{D}$ without any annotations. Our model first extracts the feature presentation $F_{\theta}(\textbf{x}_i) \in \mathbb{R}^{w\times h\times d}$, where $F_{\theta}(\cdot)$ is the backbone feature extractor,  $w$, $h$ and $d$ are the sizes of the width, height and channel of the feature map. Then the general average pooling operation is used to get the final image feature representation $\textbf{f}_i \in \mathbb{R}^{d}$ corresponding to the input image $\textbf{x}_i$. In the following, the clustering and pseudo-label refinement are all performed on these global image feature set $\mathcal{F} = \{\textbf{f}_i\}_{i=1}^{N}$.

\subsection{Pseudo-label-based Contrastive Learning}
%Following existing representative unsupervised ReID method cluster contrast~\cite{dai2021cluster}, we also simulate the pseudo-labels based on clustering results for further representation learning. Specifically, we perform DBSCAN~\cite{} clustering to group the training image feature set $\mathcal{F} = \{\textbf{f}_i\}_{i=1}^{N}$ into clustered inliers and un-clustered outliers, where the outliers are discarded in the following training epoch. Therefore, we use the cluster assignment as the pseudo identity labels. We denote the pseudo-label for the image $\textbf{x}_i$ as $\textbf{y}_i \in \mathbb{R}^{K}$, which is the one-hot encoding of the hard assignment with $K$ being the number of grouped clusters. The pseudo-labels are then used to train the deep model through the following two objectives simultaneously in the supervised manner.

To train a deep model on the dataset $\mathcal{D}$ without any annotations, we simulate the pseudo-labels based on clustering results. Specifically, we perform DBSCAN~\cite{ester1996density} clustering to group the training image features $\mathcal{F} = \{\textbf{f}_i\}_{i=1}^{N}$ into clustered inliers and un-clustered outliers, where the outliers are discarded in the following training epochs~\cite{dai2021cluster}. Therefore, we use the cluster assignment as the pseudo identity label. Let $\textbf{y}_i \in \mathbb{R}^{K}$ denote the pseudo-label for the image $\textbf{x}_i$, which is one-hot encoding of the hard assignment with $K$ being the number of grouped clusters. The pseudo-labels are then used to train the deep model through the following two objectives simultaneously in a supervised manner.

Basically, we compute the cross-entropy between the pseudo-label $\textbf{y}_i$ and network prediction $\textbf{p}_i$ as follows,
%\begin{equation}\label{LCE}
%  \mathcal{L}_{ce} = -\frac{1}{N_c}\sum_{i=1}^{N_c}\textbf{y}_i \cdot \log (\textbf{p}_i),
%\end{equation}

\begin{equation}\label{LCE}
  \mathcal{L}_{ce} = -\frac{1}{N_c}\sum_{i=1}^{N_c}\sum_{k=1}^{K}\textbf{y}_{i,k} \log (\textbf{p}_{i,k}),
\end{equation}
where $N_c$ is the number of clustered examples, $\textbf{p}_i= H_{\phi}(\textbf{f}_i) \in \mathbb{R}^{K}$ is the identity prediction vector, $H_{\phi}$ is the classifier head implemented by a fully connected layer and a softmax function, $\textbf{y}_{i,k}$ and $\textbf{p}_{i,k}$ are the $k$-th elements in $\textbf{y}_{i}$ and $\textbf{p}_{i}$, respectively.

Additionally, we also adopt the memory-bank based InfoNCE~\cite{oord2018representation,he2020momentum} loss defined as follows:
\begin{equation}\label{LCC}
  \mathcal{L}_{cc}=-\frac{1}{N_c}\sum_{i=1}^{N_c} \log\frac{\exp (\textbf{f}_i \cdot \textbf{c}_{y_i}/\tau)}{\sum_{k=1}^{K} \exp (\textbf{f}_i \cdot \textbf{c}_{k}/\tau)},
\end{equation}
where $\textbf{C}=\{\textbf{c}_1, \textbf{c}_2, \cdots, \textbf{c}_K \}$ is the cluster centers stored in the memory bank, $\textbf{c}_k \in \mathbb{R}^{d}$, $\tau$ is a temperature hyper-parameter to control the sharpness of the output, $\textbf{c}_{y_i}$ is the positive cluster center corresponding to $\textbf{f}_i$. The objective function $\mathcal{L}_{cc}$ encourages the feature vector $\textbf{f}_i$ to have a higher similarity with its corresponding positive cluster centroid $\textbf{c}_{y_i}$ and dissimilarity with other $K-1$ negative cluster centers.

During model training, the cluster memory bank $\textbf{C}\in \mathbb{R}^{K\times d}$ can be initialized with the mean feature vector of each class, where $K$ and $d$ are the number of classes and feature dimension. Specifically, the $k$-th cluster centroid $\textbf{c}_k$ can be initialized as follows,
\begin{equation}\label{CInitialization}
  \textbf{c}_k=\frac{1}{|\mathcal{F}_k|}\sum_{\textbf{f}_i\in \mathcal{F}_k}\textbf{f}_i,
\end{equation}
where $\mathcal{F}_k$ denotes the training images grouped into the $k$-th cluster in the feature space, $|\cdot|$ denotes the number of instances in the cluster set.

The cluster centroids $\textbf{C}=\{\textbf{c}_1, \textbf{c}_2, \cdots, \textbf{c}_K \}$ stored in the memory bank could be updated by the momentum updating strategy, through the feature vectors corresponding to each of their clusters during model training.
%Specifically, we adopt the mean feature vector of the instance features belonging to one cluster in a mini-batch to update the corresponding centroid, as follows,
Specifically, each cluster centroid $\textbf{c}_k$ in the memory bank can be updated as follows: $\textbf{c}_{y_i}\leftarrow \gamma \textbf{c}_{y_i} + (1-\gamma)\textbf{f}_{y_i}$,
%\begin{equation}\label{MomentumUpdating}
%  \textbf{c}_{y_i}\leftarrow \gamma \textbf{c}_{y_i} + (1-\gamma)\textbf{f}_{y_i},
%\end{equation}
where $\gamma \in [0,1]$ is the momentum updating factor, $\textbf{c}_{y_i}$ is the centroid stored in the memory bank for cluster with label $\textbf{y}_i$, $\textbf{f}_{y_i}$ is one instance feature belonging to the cluster with label $\textbf{y}_i$ in current mini-batch.

Therefore, the baseline training objective can be written as follows,
\begin{equation}\label{BaselineLoss}
  \mathcal{L} = \mathcal{L}_{cc} + \lambda \mathcal{L}_{ce},
\end{equation}
where $\lambda$ is the hyper-parameter to balance the InfoNCE and the cross-entropy loss.

In ideal conditions, the model can learn discriminative and robust representations. However, its performance is highly dependent on the quality of the generated pseudo labels $\textbf{y}_i$ in $\mathcal{L}$, which is inevitably noisy in USL person ReID task in practice.
%Therefore, we will propose a method to properly refine such noisy labels for better representation learning.
Therefore, effective measure should be taken to properly refine such noisy labels for better representation learning.

\subsection{Neighbour Consistency Guided Pseudo-Label Refinement (NCPLR)}

Complementary to the baseline training objective defined in Eq.~\ref{BaselineLoss}, the NCPLR method trains the model with refined pseudo labels through label propagation from neighbouring training examples in the feature space, to reduce the side-effects of the noisy labels on model training.

In addition to the basic one-hot pseudo-label $\textbf{y}_i$ generated by the clustering method~\cite{ester1996density}, we further propose to utilize its neighbours' predictions to refine the pseudo-label. Let $\hat{\textbf{y}} \in \mathbb{R}^{K}$  define the refined pseudo label, which is one soft target label, and can be obtained by:
\begin{equation}\label{LabelRefinement}
  \hat{\textbf{y}}_i=\alpha \textbf{y}_i + (1-\alpha)\sum_{\textbf{f}_j\in \mathcal{N}(\textbf{f}_i,\rho)}w_{ij} \cdot \textbf{p}_j,
\end{equation}
where $\textbf{p}_j= H_{\phi}(\textbf{f}_j) \in \mathbb{R}^{K}$ is the current classifier prediction for instance $\textbf{x}_j$, $\alpha$ is the hyper-parameter that controls the degree of interpolation between the original pseudo label distribution $\textbf{y}_i$ and the ensemble of its neighbours' predictions, $w_{ij}$ is the weight for each prediction $\textbf{p}_j$, which represents the importance of each neighbour's prediction contributing to the pseudo-label refinement, $\textbf{f}_j\in \mathcal{N}(\textbf{f}_i,\rho)$ means that $\textbf{f}_j$ is in the neighbour space of $\textbf{f}_i$ within radius ($<\rho$), and $\rho$ controls the size of the neighbourhood.

In order to search for the neighbourhoods $\mathcal{N}(\textbf{f}_i,\rho)$ corresponding to $\textbf{f}_i$, we need to build an overall graph of the dataset $\mathcal{D}$ in the feature space $\mathcal{F}$. Let the affinity matrix $\textbf{G}$ represent the overall graph, where $\textbf{G}_{ij}=d_{J}(\textbf{f}_i, \textbf{f}_j)\in [0,1]$ is the \emph{Jaccard Distance} between $\textbf{f}_i$ and $\textbf{f}_j$. To be specific, the $\emph{Jaccard Distance}$ can be written as,
\begin{equation}\label{JaccardDistance}
  d_{J}(\textbf{f}_i,\textbf{f}_j)=1-\frac{|\mathcal{R}(\textbf{f}_i, \kappa)\cap \mathcal{R}(\textbf{f}_j, \kappa)|}{|\mathcal{R}(\textbf{f}_i, \kappa)\cup \mathcal{R}(\textbf{f}_j, \kappa)|},
\end{equation}
where $\mathcal{R}(\textbf{f}_i, \kappa)$ is the $\kappa$-reciprocal nearest neighbors~\cite{zhong2017re}, i.e., $\mathcal{R}(\textbf{f}, \kappa)=\{\textbf{g}_i|(\textbf{g}_i\in \mathbb{N}(\textbf{f}, \kappa)) \wedge (\textbf{f}\in \mathbb{N}(\textbf{g}_i, \kappa)) \}$, and $\mathbb{N}(\textbf{f}, \kappa)$ is the $\kappa$-nearest neighbours of a probe $\textbf{f}$, $|\cdot|$ denotes the number of candidates in the set.
%Here, we adopt the $\kappa$-reciprocal nearest neighbours as they are more related and robust to probe $\textbf{f}$ than $\kappa$-nearest neighbours~\cite{}.
Afterwards, we calculate the \emph{Jaccard Distance} $d_{J}(\textbf{f}_i,\textbf{f}_j)$ between instances $\textbf{f}_i$ and $\textbf{f}_j$ by comparing their $\kappa$-reciprocal nearest neighbour set, under the assumption that \emph{ the more duplicated samples between their $\kappa$-reciprocal nearest neighbour sets, the more similar the two instances are}~\cite{zhong2017re}.

For the weighting parameter $w_{ij}$ in Eq.~\ref{LabelRefinement},
%The parameter $w_{ij}$ in Eq.~\ref{JaccardDistance} controls the importance of each neighbour's prediction to the refined pseudo-label $\hat{\textbf{y}}_i$.
the proposed method designs two types of weighting strategies: 1) the average weighting strategy, i.e., $w_{ij}=\frac{1}{|\mathcal{N}(\textbf{f}_i,\rho)|}$, where $|\mathcal{N}(\textbf{f}_i,\rho)|$ means the total number of examples within the neighborhood space of radius $\rho$ centered at the anchor point $\textbf{f}_i$; 2) the distance-based weighting strategy, which can be written as follows,
\begin{equation}~\label{PredictionWeight}
  w_{ij}=\frac{\exp(d_J(\textbf{f}_i,\textbf{f}_j)/\tau_d)}{\sum_{j=1}^{|\mathcal{N}(\textbf{f}_i,\rho)|}\exp(d_J(\textbf{f}_i,\textbf{f}_j)/\tau_d)},
\end{equation}
where $\tau_d$ is the temperature hyper-parameter. Different from some traditional neighbour consistency methods~\cite{iscen2022learning}, here we give the neighbour prediction which has larger distance from the anchor instance $\textbf{f}_i$, with relatively higher weight within the neighbourhood space of radius $\rho$, as illustrated in Eq.~\ref{PredictionWeight}.
This weighting strategy helps to improve the diversity of the refined pseudo label $\hat{\textbf{y}_i}$. When we design higher weight with closer distance, experiments show that the refined pseudo label exhibits very subtle changes, resulting in very weak label refinement.

%We have fund that the refined pseudo label shows very subtle change when given higher weight with closer distance, which results in very weak label refinement. While our proposed weighting strategy helps to improve the diversity of the refined pseudo label $\hat{\textbf{y}_i}$.

Thereafter, we can use the refined label $\hat{\textbf{y}}$ to train the neural network, and Eq.~\ref{LCE} can be re-written as follows,
%\begin{equation}\label{RefineLCE}
%  \hat{\mathcal{L}}_{ce} = -\frac{1}{N_c}\sum_{i=1}^{N_c}\hat{\textbf{y}}_i \cdot \log (\textbf{p}_i).
%\end{equation}

\begin{equation}\label{RefineLCE}
  \mathcal{L}_{ce} = -\frac{1}{N_c}\sum_{i=1}^{N_c}\sum_{k=1}^{K}\hat{\textbf{y}}_{i,k} \log (\textbf{p}_{i,k}).
\end{equation}

Please note that, in Eq.~\ref{LabelRefinement}, the feature space $\mathcal{F}$ used for constructing the overall graph $\textbf{G}$ of dataset $\mathcal{D}$, as well as for computing the weight $w_{ij}$, is extracted by previous epoch model, i.e., $T-1$ epoch model $F_{\theta}^{T-1}$.
%In Eq.~\ref{LabelRefinement}, we should note that the overall graph $\textbf{G}$ of dataset $\mathcal{D}$ is constructed based on the features extracted by the previous epoch model, i.e., $t-1$ epoch model.
Besides, the original pseudo-label $\textbf{y}_i$ is also the clustering result over the features extracted by $T-1$ epoch model $F_{\theta}^{T-1}$. Whereas, the neighbours' predictions $\textbf{p}_j$'s are the outputs of current model $F_{\theta}^{T}$. Therefore, from this point of view, our method generates the ensemble prediction $\hat{\textbf{y}}$ for each example from both the prediction of previous $T-1$ epoch model and current predictions of its neighbors. Although ensemble prediction requires a new hyper-parameter $\alpha$ and auxiliary memory to record the affinity graph $\textbf{G}$ of the dataset, it maintains a more stable and accurate prediction, especially for noisy labels in the pseudo-label set.

\subsection{Neighbour Consistency Regularization}
As we know, the estimated pseudo-labels for unsupervised learning problem inevitably contain some label noise. When learning with noisy labels, the deep neural networks can easily memorize and eventually over-fit to the noisy labels. This behavior will significantly degrade the test performances~\cite{iscen2022learning}.
% as the deep model does not generalize well~\cite{}.

Therefore, we further propose the \emph{neighbor consistency regularization}, which encourages the prediction of each example to be similar to its nearest neighbours' if they are close in the feature space. Specifically, the \emph{neighbor consistency regularization} can be denoted as follows,
 \begin{equation}\label{NCR}
   \mathcal{L}_{NCR}=\frac{1}{N_c}\sum_{i=1}^{N_c}\mathcal{KL}(\textbf{p}_i \|  \sum_{\textbf{f}_j\in \mathcal{N}(\textbf{f}_i,\rho)}\frac{1}{|\mathcal{N}(\textbf{f}_i,\rho)|}\textbf{p}_j),
 \end{equation}
where $\mathcal{KL}(\cdot || \cdot)$ is the KL-divergence to measure the difference between two distributions, $\textbf{p}_i$ and $\textbf{p}_j$ are the current classifier predictions corresponding to $\textbf{f}_i$ and $\textbf{f}_j$, respectively. The neighbourhood $\mathcal{N}(\textbf{f}_i,\rho)$ is shared with that in Eq.~\ref{LabelRefinement}. This objective function ensures that the prediction of $\textbf{x}_i$ will be consistent with its neighbours' predictions regardless of its potentially noisy label $\textbf{y}_i$, thus overcoming the over-fitting problem.

Note that the proposed \emph{neighbor consistency regularization} term $\mathcal{L}_{NCR}$ can be applied to one-stream neural network, as well as the two-stream mean-teacher framework as illustrated in Figure~\ref{overallFramework}. In our final framework, the regularization term $\mathcal{L}_{NCR}$ penalizes the divergency of each example's prediction in the teacher network from the average combination of its neighbours' predictions in the student network. Then, in Eq.~\ref{NCR}, $\textbf{p}_i$ can be marked as $\textbf{p}'_i$ representing prediction from the teacher network $F_{\theta'}$, while the second term in $\mathcal{KL}(\cdot || \cdot)$ is its neighbours' predictions from the student network $F_{\theta}$. Since the inputs for the mean-teacher networks are one example with two different augmentations, $\mathcal{L}_{NCR}$ works as self-supervision as well as neighbour consistency regularization.

Finally, the overall objective function arrives at:
\begin{equation}\label{OverallLoss}
  \mathcal{L} = \mathcal{L}_{cc} + \lambda_1 \hat{\mathcal{L}}_{ce} + \lambda_2 \mathcal{L}_{NCR},
\end{equation}
where $\lambda_1$ and $\lambda_2$ are two hyper-parameters to balance these three loss terms. The first two terms are for discriminative feature learning, while $\mathcal{L}_{NCR}$ is to overcome over-fitting problem caused by label noise in the pseudo-label set.

\section{Experiment}
\subsection{Datasets and Evaluation Protocols}
\textbf{Datasets.} We evaluate our proposed method on four large-scale person ReID datasets and one vehicle ReID dataset: Market-1501~\cite{zheng2015scalable}, DukeMTMC~\cite{zheng2017unlabeled,ristani2016performance}, MSMT17~\cite{wei2018person}, PersonX~\cite{sun2019dissecting},  VeRi-776~\cite{liu2016deep}. Market-1501 includes 32,668 images of 1,501 identities from 6 camera views. Among them, 12,936 images of 751 identities are used for training, and 19,732 images of 750 identities for testing. DukeMTMC contains 36,442 images of 1,404 identities captured from 8 cameras. It is split into 16,522 training images of 702 identities and 2,228 testing images of 702 identities. MSMT17 includes 126,441 images of 4,101 identities captured from 15 cameras. It is split into 32,621 training images of 1,041 identities and 93,820 testing images of 3,060 identities. PersonX contains 45,792 images from 1,266 identities captured from 6 cameras. It is split into 9,840 training images of 410 identities and 5,136  testing images of 856 identities. VeRi-776 contains 51,003 images of 776 vehicles captured from 20 cameras. It is split into 37,746 training images of 576 vehicles and 1,679 testing images of 200 vehicles.

\textbf{Evaluation protocol.} In testing stage, we use the mean average precision (mAP) and the cumulative match characteristic (CMC) curve to evaluate the performance of our method. The Rank-1, Rank-5, and Rank-10 accuracies are reported to represent the CMC curve, and no post-processing operation, like re-ranking, is used in our method.

\subsection{Implementation Details}
We adopt ResNet50~\cite{he2016deep} pretrained on ImageNet~\cite{deng2009imagenet} as our backbone network to conduct all the experiments. The method is implemented by PyTorch and trained on four RTX1080Ti GPUs.
%The model is initialized by the pre-trained ResNet50 model and the linear layer parameters of our model are initialized with the cluster centers.
We follow ClusterContrast~\cite{dai2021cluster} to modify the method. The size of input image is 256$\times$128 for person re-ID datasets and 224$\times$224 for vehicle ReID dataset. The batch-size is set to 256, where we randomly sample 16 identities, and each with 16 images. The momentum factor in memory bank is set to 0.1. We adopt Adam~\cite{kingma2014adam} as the optimizer, and the weight decay is 5e-4. The initial learning rate is 3.5e-4. The total epoch is 60 for all the datasets, and the learning rate is multiplied by 0.1 every 20 epochs.

At the beginning of each epoch, we perform DBSCAN~\cite{ester1996density} clustering to generate pseudo labels and obtain the distances between samples to search for the neighbors of each sample. The maximum distance in DBSCAN is set to 0.4, 0.7, 0.7, 0.7 and 0.6 for Market1501, MSMT17, PersonX, DukeMTMC and VeRi, and the maximum distance of each sample for searching neighbors is set to 0.2, 0.3, 0.2, 0.3 and 0.3 for Market1501, MSMT17, PersonX, DukeMTMC and VeRi. The temperature hyper-parameter $\tau$ and $\tau_d$ in Eq.~\ref{LCC} and Eq.~\ref{PredictionWeight} are both 0.05. During the first 50 epoch training, the temporal momentum for the teacher-student network gradually grows to 0.99, and the hyper-parameter $\lambda_2$ gradually grows to 1.0 for all the datasets.

\subsection{Experimental Results}
We compare the proposed NCPLR method with state-of-the-art USL person ReID methods on Matket-1501, DukeMTMC, MSMT17, PersonX and VeRi-776 datasets, and all the experimental results are shown in Table~\ref{table:Market-1501},~\ref{table:PersonX-sta},~\ref{table:VeRi-sta}.

The compared methods include most of recent advances on USL person ReID. Experimental results show that NCPLR is superior to state-of-the-art methods, and we obtain 86.3\% mAP and 94.3\% Rank-1 accuracy on Market1501 dataset, 74.8\% mAP and 86.6\% Rank-1 on DukeMTMC dataset, 35.7\% mAP and 66.3\% Rank-1 on MSMT17 dataset, 89.2\% mAP and 95.8\% Rank-1 on PersonX dataset, 42.0\% mAP and 85.8\% Rank-1 on Veri776 dataset. In Table~\ref{table:Market-1501},~\ref{table:PersonX-sta},~\ref{table:VeRi-sta}, $\mathcal{L}_{cc}+\mathcal{L}_{ce}$ denotes the baseline method where the model is trained by Eq.~\ref{BaselineLoss}. Our method improves the baseline method by a margin of 3.4\%, 3.1\%, 3.9\%, 2.8\% and 1.8\% in terms of mAP on Matket-1501, DukeMTMC, MSMT17, PersonX and VeRi-776 datasets, respectively.
Besides, the proposed NCPLR method outperforms the second best method by a margin of 3.2\% mAP, 2.2\% mAP, 2.1\% mAP, 4.5\% mAP and 0.4\% mAP on Matket-1501, DukeMTMC, MSMT17, PersonX and VeRi776 datasets, respectively. When compared with the most relevant and advanced pseudo-label refinement method PPLR~\cite{cho2022part}, which performs part-based pseudo-label refinement to reduce label noise, our method NCPLR surpasses PPLR~\cite{cho2022part} by a margin of 7.56\% mAP on average over the five datasets.

%, 10.9\% mAP, 4.3\% mAP, 17.4\% mAP and 0.4\% mAP on Matket-1501, DukeMTMC, MSMT17, PersonX and VeRi776 datasets, respectively.

\begin{table*}
	\footnotesize
  	\centering
  	\caption{Comparison with state-of-the-art methods on Market-1501, DukeMTMC and MSMT17 datasets under USL experimental setting.}
  	\setlength\tabcolsep{3.5pt}
	\vspace{0pt}
	\scalebox{1}[1]{
    	\begin{tabular}{c|c|cccc|cccc|cccc}
    	\toprule
    	\multirow{2}{*}{Methods} & \multirow{2}{*}{Venue} & \multicolumn{4}{c|}{Market-1501} & \multicolumn{4}{c|}{DukeMTMC} & \multicolumn{4}{c}{MSMT17} \\
    	\cline{3-14} & & mAP & Rank-1 & Rank-5 & Rank-10 & mAP & Rank-1 & Rank-5 & Rank-10 & mAP & Rank-1 & Rank-5 & Rank-10 \\
    	\hline
   %     BUC \cite{lin2019bottom}                &AAAI2019 &38.3 &66.2 &79.6 &27.5   &47.4 &62.6 &68.4 &84.5     &--   &--   &--   &--   \\
%        ECN \cite{}                             &CVPR2019 &43.0 &75.1 &87.6 &91.6   &40.4 &63.3 &75.8 &80.4     &10.2 &30.2 &41.5 &46.8 \\
%        MAR \cite{}                             &CVPR2019 &40.0 &67.7 &81.9 &87.3   &48.0 &67.1 &79.8 &84.2     &--   &--   &--   &--   \\
%        SSG \cite{}                             &ICCV2019 &58.3 &80.0 &90.0 &92.4   &53.4 &73.0 &80.6 &83.2     &13.3 &32.2 &--   &51.2 \\
%        UGA \cite{wu2019unsupervised}           &CVPR2019 &70.3 &87.2 &--   &--     &53.3 &75.0 &--   &--       &21.7 &49.5 &--   &--   \\
%        \hline
%        SoftSim \cite{lin2020unsupervised}      &CVPR2020 &37.8 &71.7 &83.8 &87.4   &28.6 &52.5 &63.5 &52.5     &--   &--   &--   &--   \\
%        TSSL \cite{wu2020tracklet}              &AAAI2020 &43.3 &71.2 &--   &--     &38.5 &62.2 &--   &--       &--   &--   &--   &--   \\
        MMCL \cite{wang2020unsupervised}        &CVPR2020 &45.5 &80.3 &89.4 &92.3   &40.2 &65.2 &75.9 &80.0     &11.2 &35.4 &44.8 &49.8 \\
%        JVTC \cite{li2020joint}                 &ECCV2020 &41.8 &72.9 &84.2 &88.7   &42.2 &67.6 &78.0 &81.6     &15.1 &39.0 &50.9 &56.8 \\
%        JVTC+\cite{li2020joint}                 &ECCV2020 &47.5 &79.5 &89.2 &91.9   &50.7 &74.6 &82.9 &85.3     &17.3 &43.1 &53.8 &59.4 \\
%        HCT \cite{zeng2020hierarchical}         &CVPR2020 &56.4 &80.0 &91.6 &95.2   &50.7 &69.6 &83.4 &87.4     &--   &--   &--   &--   \\
%        CycAs \cite{wang2020cycas}              &ECCV2020 &64.8 &84.8 &--   &--     &60.1 &77.9 &--   &--       &--   &--   &--   &--   \\
        DG-Net++ \cite{zou2020joint}                        &ECCV2020 &61.7 &82.1 &90.2 &92.7   &63.8 &78.9 &87.8 &90.4     &22.1 &48.8 &60.9 &65.9 \\
        AD-Cluster \cite{zhai2020ad}                      &CVPR2020 &68.3 &86.7 &94.4 &96.5   &54.1 &72.6 &82.5 &85.5     &--   &--   &--   &--   \\
        ECN+ \cite{zhong2020learning}                            &PAMI2020 &63.8 &84.1 &92.8 &95.4   &54.4 &74.0 &83.7 &87.4     &16.0 &42.5 &55.9 &61.5 \\
        MMT \cite{ge2020mutual}                            &ICLR2020 &71.2 &87.7 &94.9 &96.9   &65.1 &78.0 &88.8 &92.5     &23.3 &50.1 &63.9 &69.8 \\
        DCML \cite{chen2020deep}                            &ECCV2020 &72.6 &87.9 &95.0 &96.7   &63.3 &79.1 &87.2 &89.4     &--   &--   &--   &--   \\
        SPCL \cite{ge2020self}                 &NIPS2020 &73.1 &88.1 &95.1 &97.0   &65.3 &81.2 &90.3 &92.2     &19.1 &42.3 &55.6 &61.2 \\
        MEB \cite{zhai2020multiple}                             &ECCV2020 &76.0 &89.9 &96.0 &97.5   &66.1 &79.6 &88.3 &92.2     &--   &--   &--   &--   \\
        \hline
        JNTL \cite{yang2021joint}          &CVPR2021 &61.7 &83.9 &92.3 &--     &53.8 &73.8 &84.2 &--       &15.5 &35.2 &48.3 &--   \\
        JGCL \cite{chen2021joint}               &CVPR2021 &66.8 &87.3 &93.5 &95.5   &62.8 &82.9 &90.9 &93.0     &21.3 &45.7 &58.6 &64.5 \\
        IICS \cite{yang2021joint}          &CVPR2021 &72.9 &89.5 &95.2 &97.0   &59.1 &76.9 &86.1 &89.8     &18.6 &45.7 &57.7 &62.8 \\
        RLCC \cite{zhang2021refining}           &CVPR2021 &77.7 &92.8 &96.3 &97.5   &69.2 &83.2 &91.6 &93.8     &27.9 &56.5 &68.4 &73.1 \\
        MPRD \cite{ji2021meta}                  &ICCV2021 &51.1 &83.0 &91.3 &93.6   &--   &--   &--   &--       &--   &--   &--   &--   \\
        OPLG-HCD \cite{zheng2021online}   &ICCV2021 &78.1 &91.1 &96.4 &97.7 &65.6 &79.8 &88.6 &91.6 &26.9 &53.7 &65.3 &70.2 \\
        ICE \cite{chen2021ice}                  &ICCV2021 &79.5 &92.0 &97.0 &98.1   &--   &--   &--   &--       &--   &--   &--   &--   \\
        ClusterContrast~\cite{dai2021cluster}   &ArXiv2021&82.6 &93.0 &97.0 &98.1   &\underline{72.6} &\underline{84.9} &91.9 &\underline{93.9}     &27.6 &56.0 &66.8 &71.5 \\
        \hline

        SGHNG \cite{liself}                     &IJCAI2022 &75.9 &89.3 &95.9   &97.3     &65.4 &80.1 &88.9   &91.8       &24.2 &50.5 &63.0   &68.1   \\
        MCRN \cite{wu2022multi}                 &AAAI2022 &80.8 &92.5 &--   &--     &69.9 &83.5 &--   &--       &31.2 &63.6 &--   &--   \\
        SECRET~\cite{he2022secret}              &AAAI2022 &81.0 &92.6 &--   &--     &63.9 &77.9 &--   &--       &31.3 &60.4 &--   &--   \\

        HDCPD~\cite{cheng2022hybrid}            &TIP2022 &81.7 &92.4 &\underline{97.4} &98.1   &69.0 &82.9 &90.9 &93.0     &24.6  &50.2   &61.4   &65.7   \\
        MDA~\cite{ni2022meta}                   &CVPR2022 &53.0 &79.7 &-- &--   &52.4 &71.7 &-- &--     &--   &--   &--   &--   \\
         RESEL~\cite{li2022reliability}          &AAAI2022 &\underline{83.1} &\underline{93.2} &96.8   &98.0     &72.3 &83.9 &\underline{91.7}   &93.6       &\underline{33.6} &\underline{64.8} &\underline{74.6}   &\underline{79.6}   \\
        PPLR \cite{cho2022part}                 &CVPR2022 &81.5 &92.8 &97.1 &\underline{98.1}   &63.7 &79.5 &88.2 &91.3      &31.4 &61.1 &73.4 &77.8 \\
        %ISE \cite{zhang2022implicit}                             &CVPR2022 &85.3 &94.3 &98.0 &98.8   &--   &--   &--   &--       &37.0 &67.6 &77.5 &81.0 \\
   		\hline
        $\mathcal{L}_{cc}+\mathcal{L}_{ce}$  \scriptsize{( Our Baseline )}                                & --      &82.9 &92.7 &96.5 &97.4   &71.7 &84.6 &91.5 &93.7     &31.8 &62.3 &73.5 &77.8 \\
%        baseline\_refine                        & --      &85.3 &93.9 &98.0 &98.7   &73.8 &86.3 &92.3 &94.1     &33.3 &63.5 &74.1 &78.3 \\
%        baseline\_weight                        & --      &85.5 &93.6 &97.7 &98.6   &74.4 &86.3 &93.2 &94.7     &34.6 &64.4 &75.2 &79.4 \\
%        baseline\_KL                            & --      &85.7 &94.1 &97.9 &98.6   &74.4 &86.0 &93.1 &94.9     &34.6 &65.3 &75.3 &79.6 \\
        \textbf{NCPLR (Ours)} & -- &\textbf{86.3} &\textbf{94.3} &\textbf{98.0} &\textbf{98.7} &\textbf{74.8} &\textbf{86.6} &\textbf{92.6} &\textbf{94.5} &\textbf{35.7} &\textbf{66.3} &\textbf{76.9} &\textbf{80.6}\\
   		\bottomrule
    \end{tabular}}
  \label{table:Market-1501}
\end{table*}

\begin{table}[t]
\centering
\caption{Comparison with state-of-the-art methods on PersonX Dataset under USL experimental setting.}\smallskip
\label{table:PersonX-sta}
\resizebox{1.0\columnwidth}{!}{
\begin{tabular}{c|c|c|ccc }
\toprule
\textbf{Methods} &\textbf{Reference} &\textbf{mAp} &\textbf{Rank-1} &\textbf{Rank-5}  &\textbf{Rank-10} \\
\hline
MMT~\cite{ge2020mutual} &NIPS2019 &78.9 &90.6 &96.8 &98.2 \\
SPCL \cite{ge2020self}  &NIPS2020 &73.1 &88.1 &95.1 &97.0 \\
ClusterContrast~\cite{dai2021cluster}& ArXiv2021 &\underline{84.7} &\underline{94.4} &\underline{98.3} &\underline{99.3} \\
%HDCPD~\cite{cheng2022hybrid}   &TIP2022 &84.1 &94.4 &98.7 &99.5 \\
PPLR~\cite{cho2022part}         &CVPR2022 &71.8 &89.4 &96.2 &98.4 \\
\hline
 $\mathcal{L}_{cc}+\mathcal{L}_{ce}$ (Our Baseline) & --  &86.4 &94.9 &98.5 &99.3 \\
%baseline\_refine & --  &88.6 &95.9 &98.9 &99.5 \\
%baseline\_weight & --  &88.8 &95.7 &99.0 &99.6 \\
%baseline\_KL & --  &88.8 &95.5 &98.8 &99.5 \\
\textbf{NCPLR (Ours)} & -- &\textbf{89.2} &\textbf{95.8} &\textbf{98.9} &\textbf{99.5}\\
\bottomrule
\end{tabular}}
\end{table}

\begin{table}[t]
\centering
\caption{Comparison with state-of-the-art methods on VeRi-776 Dataset under USL experimental setting.}\smallskip
\label{table:VeRi-sta}
\resizebox{1.0\columnwidth}{!}{
\begin{tabular}{c|c|c|ccc }
\toprule
%\hline
%    \multirow{1}{*}{Methods} & {\multirow{1}{*}{Feature}}& \multicolumn{2}{c|}{$Rank 1@$} &\multicolumn{2}{c}{$Rank 5@$}  \\
%    \cline{3-6}
\textbf{Methods} &\textbf{Reference} &\textbf{mAp} &\textbf{Rank-1} &\textbf{Rank-5}  &\textbf{Rank-10} \\
\hline
MMT~\cite{ge2020mutual} &NIPS2019 &35.5 &74.6 &82.6 &87.0 \\
SPCL \cite{ge2020self}  &NIPS2020 &38.9 &80.4 &86.8 &89.6 \\
ClusterContrast~\cite{dai2021cluster}   &ArXiv2021 &40.3 &84.6 &89.2 &91.6 \\
RLCC~\cite{zhang2021refining}   &CVPR2021 &39.6 &83.4 &88.8 &90.9 \\
PPLR~\cite{cho2022part}         &CVPR2022 &\underline{41.6} &\underline{85.6} &\textbf{91.1} &\textbf{93.4} \\
\hline
 $\mathcal{L}_{cc}+\mathcal{L}_{ce}$ (Our Baseline) & --  &40.2 &83.0 &86.8 &90.4 \\
%baseline\_refine & --  &41.4 &84.6 &89.2 &82.0 \\
%baseline\_weight & --  &41.5 &86.1 &89.8 &92.4 \\
%baseline\_KL & --  &41.6 &86.4 &90.3 &92.6 \\
\textbf{NCPLR (Ours)} & -- &\textbf{42.0} &\textbf{85.8} &\underline{90.5} &\underline{92.6}\\
\bottomrule
\end{tabular}}
\end{table}

\begin{table*}
	\footnotesize
  	\centering
  	\caption{Ablation study on Market-1501, DukeMTMC, MSMT17, PersonX and VeRi-776 datasets.}
  	\setlength\tabcolsep{6pt}
	\vspace{0pt}
	%\resizebox{2.0\columnwidth}{!}{
	\resizebox{2.0\columnwidth}{!}{
    	\begin{tabular}{c|cc|cc|cc|cc|cc}
    	\toprule
    	\multirow{2}{*}{Methods}  &\multicolumn{2}{c|}{Market-1501} &\multicolumn{2}{c|}{DukeMTMC} &\multicolumn{2}{c|}{MSMT17} &\multicolumn{2}{c|}{PersonX} &\multicolumn{2}{c}{VeRi-776}\\
    	\cline{2-11}  & mAP & Rank-1 & mAP & Rank-1 & mAP & Rank-1 & mAP & Rank-1 & mAP & Rank-1 \\
    	
   		\hline
        $\mathcal{L}_{cc}$                                                      &80.8 &91.7   &71.6 &84.3   &29.2 &57.6     &84.5 &94.5     &39.8 &83.2 \\
        $\mathcal{L}_{cc}+\mathcal{L}_{ce}$ (Our Baseline)                      &82.9 &92.7   &71.7 &84.6   &31.8 &62.3     &86.4 &94.9     &40.2 &83.0 \\
        $\mathcal{L}_{cc}+\hat{\mathcal{L}}_{ce}$(w/o weight)                   &85.3 &93.9   &73.8 &86.3   &33.3 &63.5     &88.6 &\textbf{95.9}     &41.4 &84.6 \\
        $\mathcal{L}_{cc}+\hat{\mathcal{L}}_{ce}^w$( weight)                    &85.5 &94.2   &74.4 &86.3   &34.6 &64.4     &88.8 &95.7     &41.5 &86.1 \\
        $\mathcal{L}_{cc}+\hat{\mathcal{L}}_{ce}^w+\mathcal{L}_{NCR}^s$         &85.7 &94.1   &74.5 &86.0   &34.6 &65.3     &88.8 &95.5     &41.6 &\textbf{86.4} \\
        %$\mathcal{L}_{cc}+\hat{\mathcal{L}}_{ce}^w+\mathcal{L}_{NCR}^{st-self}$ &85.9 &94.0   &74.1 &86.0   &33.8 &63.6     &88.8 &96.0     &41.1 &83.8 \\
%       baseline\_KL                                &85.7 &94.1 &97.9 &98.6   &74.4 &86.0 &93.1 &94.9     &34.6 &65.3 &75.3 &79.6 &31.8 &62.3 &73.5 &77.8\\
        \textbf{NCPLR (Ours)} &\textbf{86.3} &\textbf{94.3} &\textbf{74.8} &\textbf{86.6}  &\textbf{35.7} &\textbf{66.3}  &\textbf{89.2} &95.8 &\textbf{42.0} &85.8 \\
   		\bottomrule
    \end{tabular}}
  \label{table:Ablation}
\end{table*}

\subsection{Ablation Study}
The proposed NCPLR training framework contains three items as shown in Eq.~\ref{OverallLoss}: 1) the memory-bank based InfoNCE objective $\mathcal{L}_{cc}$; 2) the cross-entropy loss with refined pseudo labels $\hat{\mathcal{L}}_{ce}$; 3) the \emph{neighbour consistency regularization} $\mathcal{L}_{NCR}$ to overcome over-fitting, as well as working as self-supervision. To reveal how each ingredient contributes to performance improvement, we conduct comprehensive ablation study to analyze different elements in Eq.~\ref{OverallLoss}.

Specifically, we implement six variants of the proposed method as shown in Table~\ref{table:Ablation}: 1) $\mathcal{L}_{cc}$: Using only the memory-based InfoNEC loss to train the network as one baseline; 2)$\mathcal{L}_{cc}+\mathcal{L}_{ce}$: Training the network by $\mathcal{L}_{cc}$ and the cross-entropy loss $\mathcal{L}_{ce}$ with clustering-based pseudo-labels jointly; 3) $\mathcal{L}_{cc}+\hat{\mathcal{L}}_{ce}$(w/o weight): Training the network using $\mathcal{L}_{cc}+\hat{\mathcal{L}}_{ce}$ with refined pseudo labels as Eq.~\ref{LabelRefinement}, where $w_{ij}=\frac{1}{|\mathcal{N}(\textbf{f}_i,\rho)|}$; 4) $\mathcal{L}_{cc}+\hat{\mathcal{L}}_{ce}^w$( weight): Training the network using $\mathcal{L}_{cc}+\hat{\mathcal{L}}_{ce}$ with refined pseudo labels as Eq.~\ref{LabelRefinement}, where the weighting strategy is set as Eq.~\ref{PredictionWeight}; 5) $\mathcal{L}_{cc}+\hat{\mathcal{L}}_{ce}^w+\mathcal{L}_{NCR}^s$: Additionally applying the \emph{neighbour consistency regularization} $\mathcal{L}_{NCR}$ to one stream neural network; %6)$\mathcal{L}_{cc}+\hat{\mathcal{L}}_{ce}^w+\mathcal{L}_{NCR}^{st-self}$: Applying $\mathcal{L}_{NCR}$ to two-stream teacher-student networks, where we just perform self-supervision of the predictions for one example with two different data augmentations;
6) NCPLR: Our final training objective as defined in Eq.~\ref{OverallLoss}, which applies $\mathcal{L}_{NCR}$ to two-stream teacher-student networks.

The performances of these method variants are summarized in Table~\ref{table:Ablation}, where we have done ablation experiments on all the five datasets. By comparing the performances of methods $\mathcal{L}_{cc}+\mathcal{L}_{ce}$ and $\mathcal{L}_{cc}+\hat{\mathcal{L}}_{ce}$(w/o weight), we can clearly see that the proposed neighbour consistency guided pseudo-label refinement strategy can greatly improve the model performances by an average margin of 1.48\% mAP over the baseline method $\mathcal{L}_{cc}+\mathcal{L}_{ce}$, on these five datasets. When additional weighting strategy is used as Eq.~\ref{PredictionWeight}, another average 0.48\% mAP can be obtained on average. Specifically, comparing the methods  NCPLR and $\mathcal{L}_{cc}+\hat{\mathcal{L}}_{ce}^w$( weight), we can clearly see that $\mathcal{L}_{NCR}$ helps to improve the performance by an average margin of 0.64\% mAP on these five datasets, when applied to the teacher-student networks.

\subsection{Hyper-parameter Analysis}
We analyze the impact of four important hyper-parameters in the proposed method: 1) the parameter $\alpha$ in Eq.~\ref{LabelRefinement} to control the degree of interpolation between the original pseudo label $\textbf{y}_i$ and the ensemble of neighbours' predictions; 2) the two hyper-parameters $\lambda_1$ and $\lambda_2$ in Eq.~\ref{OverallLoss} to balance the three items, i.e., $\mathcal{L}_{cc}$, $\mathcal{L}_{ce}$ and $\mathcal{L}_{NCR}$; 3) the parameter of radius $\rho$ in Eq.\ref{LabelRefinement} and Eq.~\ref{NCR} (i.e., $\textbf{f}_j\in \mathcal{N}(\textbf{f}_i,\rho)$) to control the size of the neighborhood. To keep pace with our experimental exploration process, we first analyze these two parameters $\alpha$ and $\lambda_1$, and then tune the values of $\lambda_2$ and $\rho$ while keeping the others fixed.

\begin{figure}
    \centering
     \subfloat[\quad $\alpha$]{
         \label{fig:alpha}
         \includegraphics[width=0.46\linewidth,height=0.48\linewidth]{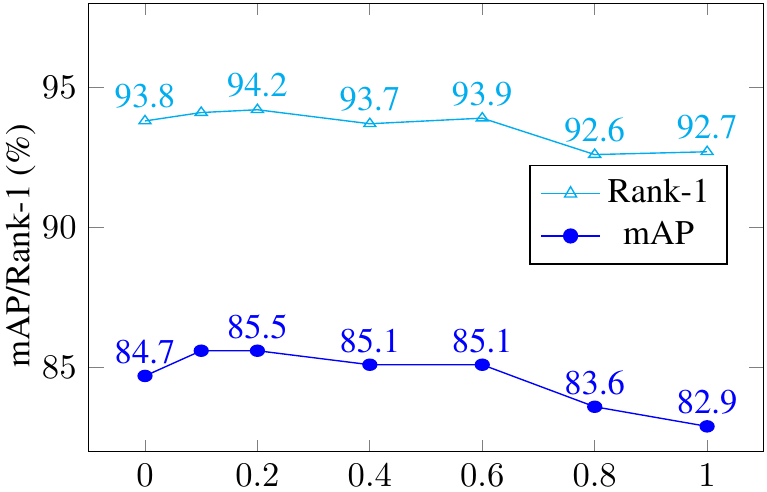}
     }
%     \hspace{-0.3cm}
     \subfloat[\quad $\lambda_1$]{
        \label{fig:lambda1}
         \includegraphics[width=0.46\linewidth,height=0.48\linewidth]{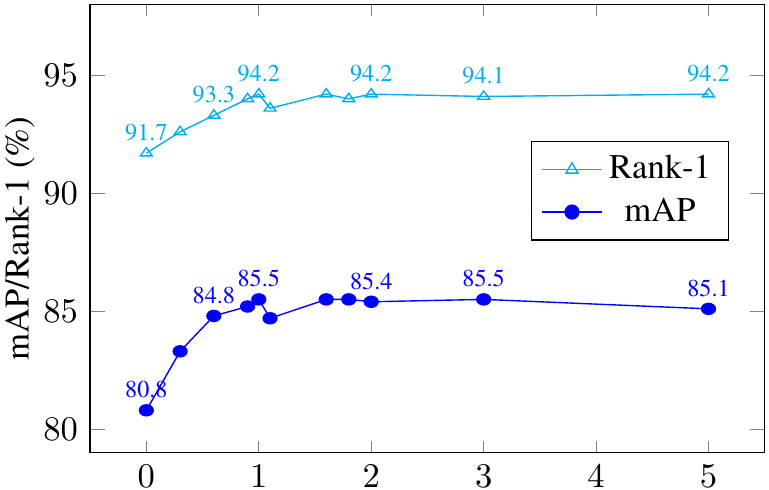}
    }

  \caption{Parameter analysis of (a) $\alpha$ and (b) $\lambda_1$ on Market1501.  }
  \label{figure_alpha_lambda1}
\end{figure}

% \begin{figure*}[htbp]
%     %\centering
%      \subfloat[]{
%          \label{fig:small_orig1}
%          \includegraphics[width=0.24\linewidth,height=0.25\linewidth]{alpha.pdf}
%      }
%      \hfill
% %     \hspace{-0.3mm}
%      \subfloat[ ]{
%         \label{fig:small_zsclip1}
%          \includegraphics[width=0.24\linewidth,height=0.25\linewidth]{lambda1.pdf}
%     }

%     \hfill
%  %   \hspace{-0.3mm}
%      \subfloat[ ]{
%          \label{fig:small_orig2}
%          \includegraphics[width=0.24\linewidth,height=0.25\linewidth]{lambda2.pdf}
%      }
%      \hfill
%  %    \hspace{-0.3mm}
%      \subfloat[]{
%         \label{fig:small_zsclip2}
%          \includegraphics[width=0.24\linewidth,height=0.25\linewidth]{rho.pdf}
%     }

%      % \includegraphics[width=1\columnwidth]{visualize_small4x3noname.pdf} % Reduce the figure size so that it is slightly narrower than the column. Don't use precise values for figure width.This setup will avoid overfull boxes.
%   \caption{Visualization of the attention map of the image encoder. (a) Original Image. (b)Zero-Shot CLIP/CoOp. (c)Zero-Shot CLIP/CoOp. (d)Zero-Shot CLIP/CoOp. }
%   \label{figure_vis1}
% \end{figure*}

\begin{figure}
    \centering
     \subfloat[\quad $\lambda_2$]{
         \label{fig:lambda2}
         \includegraphics[width=0.46\linewidth,height=0.48\linewidth]{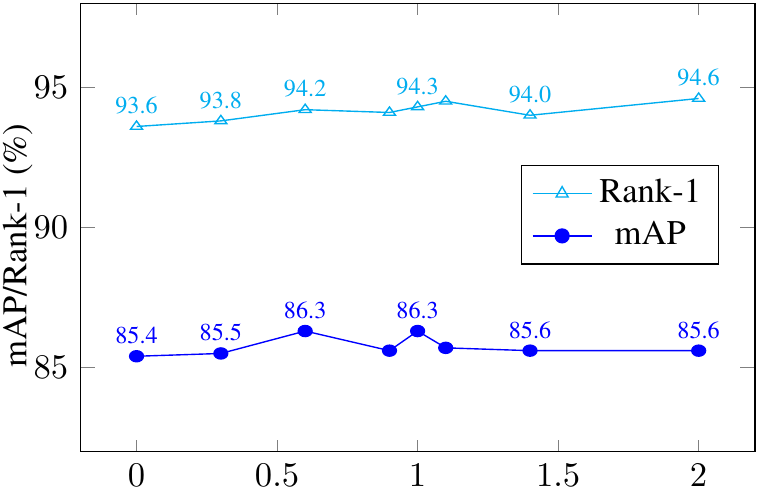}
     }
   %  \hspace{-0.1cm}
     \subfloat[\quad  $\rho$]{
        \label{fig:rho}
         \includegraphics[width=0.46\linewidth,height=0.48\linewidth]{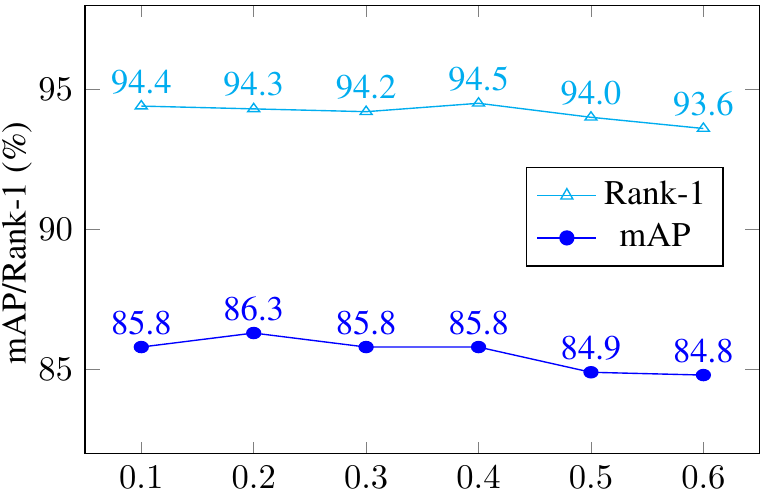}
    }

  \caption{Parameter analysis of (a) $\lambda_2$ and (b) $\rho$ on Market1501. }
  \label{figure_lambda_rho}
\end{figure}

To investigate the effect of hyper-parameters on the model performances, we conduct comprehensive experiments with various values of these parameters. All the experimental analysis results in terms of Rank-1 accuracy and mAP with varying values are shown in Figure~\ref{fig:alpha},~\ref{fig:lambda1},~\ref{fig:lambda2},~\ref{fig:rho}.
%As shown in Figure~\ref{overallFramework}, we conduct comprehensive experiments with various values of $\alpha$ to investigate its effect on the model performance, while keeping $\lambda_1=1.0$ and $\rho=0.2$.
Figure~\ref{fig:alpha} illustrates the model performances with varying values of $\alpha$ from 0.0 to 1.0, when $\lambda_1=1.0$ and $\rho=0.2$ under the $\mathcal{L}_{cc}+\hat{\mathcal{L}}_{ce}^w$ experiment setting. When $\alpha =0$, it means that the refined label $\hat{\textbf{y}}_i$ is just from current neighbors' predictions, which could obtain 84.7\% mAP and 93.8 Rank-1 accuracy. When $\alpha=1.0$, it means that the refined label $\hat{\textbf{y}}_i$ is the original clustering results $\textbf{y}_i$, whose results are 82.9\% mAP and 92.7\% Rank-1 Accuracy. Clearly, we can see that the model can obtain best performances when $\alpha=0.2$, indicating that the combination of original clustering result and current neighbours' predictions helps to improve the model performances. Figure~\ref{fig:lambda1} illustrates the model performances with varying values of $\lambda_1$ from 0.0 to 5.0, when $\alpha=0.2$ and $\rho=0.2$ under the same experiment setting. We can see that the model obtains best performance when $\lambda_1=1.0$. Figure~\ref{fig:lambda2} illustrates the model performances with varying values of $\lambda_2$ from 0.0 to 2.0, when $\lambda_1=1.0,\alpha=0.2$ and $\rho=0.2$ under the fully NCPLR experiment setting. We can see that the model obtains best performances when $\lambda_2=1.0$. According, we also conduct experiment analysis on $\rho$ with varying values from 0.1 to 0.6 as shown in Figure~\ref{fig:rho}, under the same experiment setting. We can see that the model obtains best performances when $\rho=0.2$. Based on these experimental results, we set $\alpha=0.2$, $\lambda_1=1.0$, $\lambda_2=1.0$ and $\rho=0.2$.

\section{Conclusion and Limitation}

This paper proposes a neighbour consistency guided pseudo-label refinement (NCPLR) framework for USL person ReID. The proposed NCPLR can be regarded as a transductive form of label propagation for pseudo-label refinement. The refined label for each training example can be obtained by the original clustering result and a weighted combination of its neighbours' predictions,
with weights determined by the similarities in the feature space.
Besides, we consider the clustering-based USL person ReID task as a label-noise learning problem, and propose the \emph{neighbour consistency regularization} to reduce model over-fitting to noisy labels. The proposed NCPLR algorithm is simple yet effective, and can be seamlessly integrated into existing clustering-based USL person ReID methods. In the future, we will apply the proposed method to other USL tasks.

\vspace{1mm}
\noindent \textbf{Limitation.}
One major limitation in this study is that, the success of the proposed NCPLR strategy relies on the assumption that the percentage of incorrect labels in the pseudo-label set is relatively small so as not to overwhelm the correct labels. If the clustering results are very poor, the effectiveness of such pseudo-label refinement strategy would be undermined.

% One major limitation in this study is that we just adopt one fully-connected layer with $ K\times K$ output nodes to work as the transition neural network (TNN) for learning $T(\textbf{x})$, which is a little bit simple. In the future, we will make deep analysis on the TNN design theoretically and practically, to learn robust classifier on the noisy dataset.
%%%%%%%%% REFERENCES
{\small
\bibliographystyle{ieee_fullname}
\bibliography{egbib}
}

\end{document}